\documentclass[]{ailab}

\usepackage{amsmath,amssymb}
\usepackage{graphicx}
\graphicspath{{figures/}{.}}
\usepackage{booktabs}
\usepackage{longtable}
\usepackage{natbib}
\usepackage{xspace}
\usepackage{float} % for [H] placement of key figures
\setcitestyle{square,comma,numbers,sort&compress}

\providecommand{\ours}{\textsc{Long-Horizon-Terminal-Bench}\xspace}

\title{\centering \ours{}: Testing the Limits of Agents on Long-Horizon Terminal Tasks with Dense Reward-Based Grading}

\author[1,2]{Zongxia Li\textsuperscript{$\dagger$}}
\author[1,3]{Zhongzhi Li\textsuperscript{$\dagger$}}
\author[1]{Yucheng Shi\textsuperscript{$\dagger$}}
\author[1,4]{Ruhan Wang}
\author[1,5]{Junyao Yang}
\author[2]{Zhichao Liu}
\author[2]{Xiyang Wu}
\author[6]{Anhao Li}
\author[7]{Chengsong Huang}
\author[8]{Ergan Shang}
\author[4]{Yue Yu}
\author[9]{Ninghao Liu}
\author[10]{Lichao Sun}
\author[1]{Haitao Mi}
\author[1]{LeoweiLiang}

\affiliation[1]{Tencent HY LLM Frontier}
\affiliation[2]{University of Maryland, College Park}
\affiliation[3]{University of Georgia}
\affiliation[4]{Indiana University}
\affiliation[5]{National University of Singapore}
\affiliation[6]{University of Minnesota, Twin Cities}
\affiliation[7]{Washington University in St. Louis}
\affiliation[8]{Carnegie Mellon University}
\affiliation[9]{The Hong Kong Polytechnic University}
\affiliation[10]{Lehigh University}

\email{zli12321@umd.edu, zongxiali@global.tencent.com, zhongzhili@global.tencent.com}

\contribution{$^\dagger$ Core contribution.}

\abstract{
AI agents are capable of autonomously completing short, well-specified tasks with high efforts to keep high success rates. 
Existing terminal benchmarks focus on trivial problems that finish within a few minutes (e.g., making two chess moves or reconstructing a corrupted image) and comes with over-simplified graders only for the final outcome. 
This over-simplifies agent evaluation and ignores intermediate progress and partial solutions, leading to sparse reward signals and biased measurements of agent capability.

We introduce \ours{}, a difficult terminal benchmark of 46 long-horizon tasks spanning nine categories, including e.g, interactive games, experiment reproduction, software engineering, multimodal analysis, and scientific computing. 
Each task follows the terminal-bench style setup with a reference solution or simulation engine and is decomposed into more fine-grained graded subtasks. The grader is designed to give dense intermediate rewards and partial credit on even uncompleted tasks, so agents are evaluated not only on whether they reach the final goal, but also on how far they progress.

Tasks typically require hundreds of episodes and more than 30 minutes to hours of execution time, stressing planning, long-context management, and iterative bug refinement rather than one-shot solutions. 
We evaluate 17 frontier models under a shared terminal agent, and find that agents consume on average 9.8M tokens per task, with around 239 episodes and 88.9 minutes of execution time, an order of magnitude more demanding than prior terminal-based benchmarks such as terminal-bench 2, with an average execution time of 20 to 30 minutes and 20 to 30 episodes per task. 
The strongest frontier model we evaluated, Grok 4.5, achieves only $28.3\%$ pass rate at 0.95 partial reward threshold and $19.6\%$ at 1.0 reward threshold, while the mean pass rate across models is still only $6.4\%$ at partial reward threshold 0.95 and $3.2\%$ at 1.0 reward threshold, showing substantial headroom for improvement. 
We qualitatively analyze common failure modes and error patterns, and release \ours{} to evaluate progress on robust, long-horizon terminal agents.
}

\abstract{
AI agents have become increasingly capable of autonomously completing short, well-specified tasks. However, existing terminal benchmarks largely focus on relatively simple problems that finish within a few minutes and are typically evaluated only by their final outcome. This setup overlooks intermediate progress and partial solutions, leading to sparse reward signals and an incomplete picture of agent capability.

We introduce \ours{}, a challenging terminal benchmark of 46 long-horizon tasks spanning nine categories, including experiment reproduction, software engineering, multimodal analysis, interactive games, and scientific computing. Each task follows a Terminal-Bench-style setup with a reference solution or simulation engine, but is further decomposed into fine-grained graded subtasks. 
This design enables dense intermediate rewards and partial credit, allowing evaluation to capture not only whether an agent reaches the final goal, but also how far it progresses on difficult, open-ended workflows.

Tasks in \ours{} typically require hundreds of episodes and tens of minutes to hours of execution, stressing long-horizon planning, long-context management, and iterative debugging rather than one-shot problem solving. 
We evaluate 17 frontier models and find that agents consume on average 9.8M tokens per task, with roughly 239 episodes and 88.9 minutes of execution time per run, making \ours{} substantially more demanding than prior terminal-based benchmarks. 
Even the strongest tested model, Grok 4.5, achieves only $28.3\%$ pass@1 at a partial-reward threshold of 0.95 and $19.6\%$ at a perfect-reward threshold of 1.0, while the mean pass rate across models is just $6.4\%$ and $3.2\%$ under the two thresholds, respectively. These results reveal substantial headroom for improvement. We further analyze common failure modes and error patterns, and release \ours{} to support future progress on robust long-horizon terminal agents.
}

\headercontent{{\sffamily\bfseries Project page:} \url{https://zli12321.github.io/LHTB/}}

\date{\today}

\begin{document}
\thispagestyle{firstheader}
\maketitle

\section{Introduction}

AI agents built on large language models (LLMs) are rapidly improving at autonomous decision making and tool use~\cite{du2026survey, chowa2026language}.
Recent work has demonstrated impressive performance on short, well-scoped tasks such as fixing a code repository issue, completing a coding ticket, or issuing a handful of shell commands~\cite{merrill2026terminalbenchbenchmarkingagentshard}.
However, they still cover only a narrow slice of the long-horizon, domain-specific, and practically important workflows that human experts care about in practice~\citep{zhou2023webarena,xie2024osworld,merrill2026terminalbenchbenchmarkingagentshard}.

Nowadays, many real workflows are \emph{long-horizon}: they require agents to execute hundreds of steps, maintain and update plans over tens of minutes to hours, and manage evolving long-context state~\cite{wang2026long}.
Examples include reproducing results from published research papers~\cite{song2026drclaw}, installing environments from a github repo, auditing complex multimodal datasets, debugging compiler toolchains, or shepherding a multi-stage ML training pipeline.
In these settings, agents must repeatedly use terminal command lines to read and write files, run scripts, inspect partial outputs, revise their plans, and recover from mistakes.
The outcome is determined not by a single action, but by sustained, coherent progress over a long sequence of decisions~\cite{liu2026klong}.

Existing benchmarks lack fine-grained verifiers to evaluate agents on long-horizon terminal tasks~\cite{jimenez2024swebench}.
Dominant terminal and software-engineering benchmarks typically evaluate within a few minutes to at most an hour, and grade agents solely on whether the final state satisfies a test suite~\cite{desai2026swemarathon,frontierswe2026}.
This leaves two gaps.
First, short tasks or narrow horizons understate the difficulty of real workflows that require long-horizon navigation, errored step revision, and multi-stage debugging, not just local patching~\cite{merrill2026terminalbenchbenchmarkingagentshard, feng2026longcli}.
Second, outcome-only grading yields extremely sparse reward signals: an agent that completes most intermediate steps but fails at the final step can receive the same score as one that fails from the beginning~\cite{lin2025cuarewardbench}.
Both issues make it difficult to understand how well current agents can sustain progress on long workflows, or where they fail.

In this paper we introduce \emph{\ours{}}, a benchmark designed explicitly for long-horizon terminal tasks.
\ours{} consists of 46 tasks spanning nine categories, including experiment reproduction, interactive games, software engineering, multimodal analysis, and scientific computing.
Each task is hosted in a containerized terminal environment and is decomposed into graded subtasks with intermediate checks.
Rather than asking agents to \texttt{solve everything or fail}, the grader assigns credit for partial progress toward a final goal, providing dense feedback along long trajectories while preserving the realism of everyday tasks.

\ours{} targets domains where the completion of the final big goal depends on the optimal and completion of the sub goals, and agents must execute \emph{many} steps over \emph{long} time horizons,.
Across 17 frontier agents evaluated, rollouts on \ours{} average 239 episodes, 9.8M tokens, and 88.9 minutes of wall-clock time before exit under a 90-minute timeout.
These numbers are substantially larger than prior terminal and code-editing benchmarks such as Terminal-Bench 2 tasks and SWE-Bench.
Yet even in this relaxed grading, current agents struggle: our strongest reported configuration, Grok 4.5, achieves only 28.3\% success rate at a 0.95 reward threshold, and the average pass rate across models is 6.4\%.
This suggests that long-horizon execution remains a central bottleneck for current systems.

We also conduct qualitative and quantitative analyses of agent failures on \ours{}.
Our analysis shows that current agents often fail not because every local step is wrong, but because they cannot reliably sustain progress, verify completion, and finish long-horizon tasks within budget.
Dense rewards make these differences visible by separating timeout-driven incomplete progress from premature stopping and weak self-verification.
We release \ours{} and the accompanying evaluation harness to support future work on agents that can plan, verify, and execute reliably over long horizons.
\section{\ours{}}
\label{sec:background}

\ours{} builds on the Terminal-Bench formulation~\cite{merrill2026terminalbenchbenchmarkingagentshard}, where each task is a containerized terminal environment with a natural-language instruction, all relevant files and tools, and a reference solution that can be executed entirely from the command line.
Tasks span diverse domains and are explicitly designed so that progress requires \emph{continued} work over many steps: reading and modifying code and data, running long scripts, inspecting partial outputs, and iteratively debugging.

Unlike Terminal-Bench, which evaluates tasks in a binary solved/unsolved fashion, \ours{} is graded through a subtask-based scheme.
Each task is decomposed into a small set of semantically meaningful subtasks with their own checks, and agents are scored by their completion rate across these subtasks.
This design provides partial credit and a much denser signal about long-horizon performance, revealing how far agents get through complex workflows rather than only whether they reach the final goal.

% Introduce any necessary background, notation, or benchmark details.
% Define the setting clearly (inputs, outputs, objectives, constraints).

\subsection{Task formulation}

A \ours{} task follows the Terminal-Bench formulation~\cite{merrill2026terminalbenchbenchmarkingagentshard}: it is specified as a Harbor task with (1) a natural-language instruction, (2) a Docker image that defines the terminal environment, (3) a task configuration file, and (4) an oracle implementation or simulator used for grading.
The instruction describes the overall long-horizon goal (e.g., reproduce a figure, repair a robotics SLAM pipeline, audit a multimodal dataset) and is the only specification the agent sees.
The Docker image contains all relevant assets, code, data, tools, and helper scripts, so that an agent can complete the task entirely from the terminal.
Tasks are interactive: after the container starts, the agent issues shell commands, edits files, runs scripts, and inspects intermediate outputs over hundreds of steps until it either succeeds or times out.

While the task format matches Terminal-Bench 2, \ours{} is explicitly designed so that a single subtask typically requires many reasoning steps and many minutes to hours of work and dozens to hundreds of distinct operations.
Long-horizon structure comes both from the underlying domains (multi-stage ML pipelines, campaign-style games, scientific audits) and from the way we factor goals into a sequence of intermediate targets that must be discovered and executed over time.

\begin{figure}[H]
  \centering
  \includegraphics[width=1.0\linewidth]{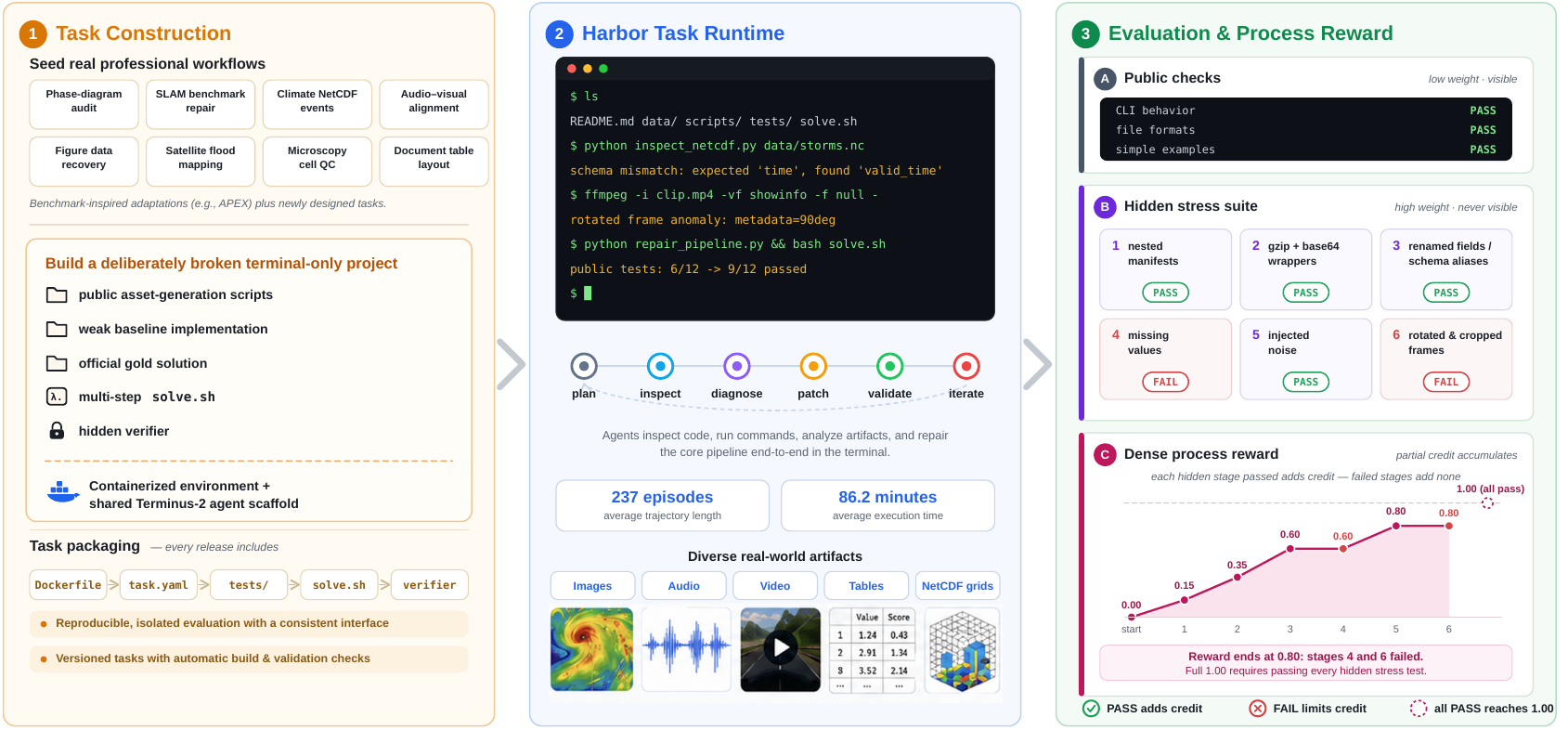}
  \caption{The overall structure of long-horizon terminal task with dense reward grading. Partial rewards show how far models are into completing the task.}
  \label{fig:pipeline}
\end{figure}

\subsection{Subtask-based grading}

Each \ours{} task is graded by a deterministic grader that runs \emph{inside} the container at the end of a rollout.
Instead of returning a strict single pass/fail, the grader computes scores for a small set of subtasks
$\{s_1,\dots,s_K\}$ that together describe the intended workflow.
For every subtask $s_k$ it outputs a normalized score $r_k \in [0,1]$; the task reward is

\[
R \;=\; \frac{\sum_{k=1}^K w_k\, r_k}{\sum_{k=1}^K w_k},
\]

where $w_k$ are non‑negative weights (by default all equal, with higher weight on the final goal when needed).
Subtasks are chosen to correspond to meaningful intermediate goals in the workflow, and each is evaluated using objective evidence from the final container state, such as files, outputs, test results, or simulator responses.:

- \textbf{Binary subtasks.}
Checks are implemented as strict Boolean conditions over the environment state, such as whether all unit tests exit successfully, whether a service responds on the expected port, or whether required experiment scripts run without error.
These subtasks return $r_k \in \{0,1\}$ depending on whether the corresponding programmatic check passes.

- \textbf{Continuous or thresholded subtasks.}
  For quantitative targets (e.g., reproducing a figure, matching a metric, or achieving a given speedup) we use continuous scores.
  Examples include \textit{1.0 if the reproduced metric is within tolerance of the reference, decreasing linearly to 0 as error grows}, or \textit{fraction of held‑out examples on which the model’s predictions match the oracle}.
  This yields meaningful partial credit when an agent gets close but not perfect.

- \textbf{Episode-aggregating subtasks.}
  Campaign‑style tasks (e.g., games or repeated audits) aggregate across episodes.
  A typical subtask is the fraction of episodes or levels in which the environment’s internal success flag is triggered, or the mean normalized reward reported by the simulator.
  This measures whether an agent can perform a long-horizon behavior reliably across many episodes, rather than succeeding only once by chance.

The overall task reward $R$ is used for all reported metrics.
A task is counted as \emph{resolved} for sucess if $R \geq \tau$\footnote{we use a relaxed threshold such as $\tau = 0.95$}.
We also report mean $R$ across tasks to capture how far agents progress on problems they cannot fully solve.
% Because the grader records the full vector $(r_1,\dots,r_K)$, \ours{} enables fine‑grained analysis of where long‑horizon workflows break down: some agents reliably clear early setup subtasks but fail at later optimization or verification stages, while others struggle to make any progress past initialization.

\begin{figure}[H]
  \centering
  \includegraphics[width=0.78\linewidth]{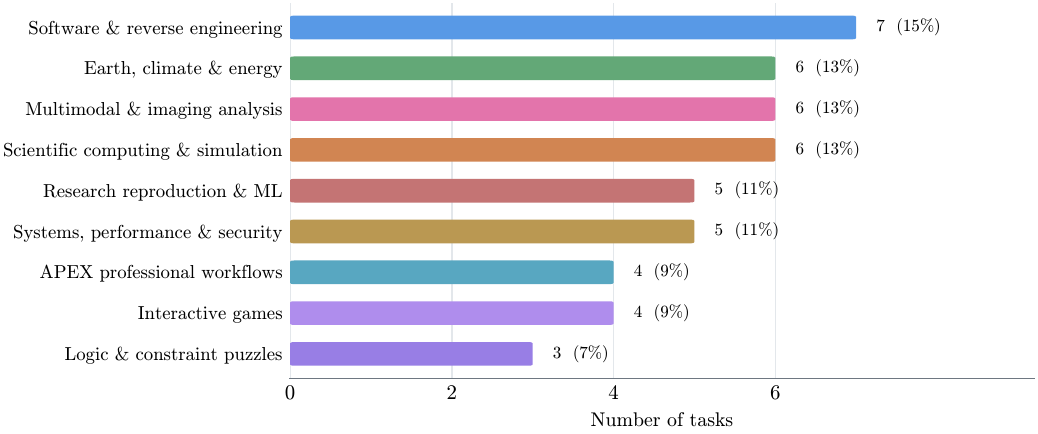}
  \caption{\textbf{Task distribution across the LHTB categories.}
  The 46 tasks span nine long-horizon domains, ranging from software and reverse engineering to scientific computing, earth and climate science, and multimodal analysis.}
  \label{fig:lhtb-categories}
\end{figure}

\subsection{Dataset construction}

We construct \ours{} by instantiating a common recipe across a diverse set of real-world long-horizon workflows~\cite{vidgen2026apex, wu2026co}.
Each task begins with a realistic professional data-processing or engineering problem, such as materials phase-diagram auditing, robotics SLAM benchmark repair, climate NetCDF extreme-event detection, audio--visual event alignment, scientific-figure data reconstruction, satellite flood change detection, microscopy cell-count quality control, or scanned-document table reconstruction.
Some of these professional-workflow tasks are adapted from or inspired by prior long-horizon agent benchmarks such as APEX-Agents~\cite{vidgen2026apex}, while others are newly created for \ours{}.

For each problem, we build a complete but deliberately broken terminal-only project.
Agents must work entirely through the terminal: inspecting code, running commands, and analyzing artifacts such as images, audio, video, tables, and NetCDF files in order to diagnose and repair the core pipeline.
Each task includes reproducible public asset-generation scripts, a weak baseline implementation, an official gold solution, a multi-step \texttt{solve.sh}, and a hidden verifier.

To discourage overfitting to superficial public tests, the public checks only validate command-line behavior, file formats, and a small number of simple examples, and they carry relatively low reward weight.
Most of the reward is assigned through hidden stress cases that dynamically generate harder inputs and schema variations, such as nested manifests, gzip-plus-base64 wrappers, renamed fields, missing values, injected noise, rotated or cropped images, anomalous frames, and alternative coordinate or time-dimension conventions.
As a result, agents cannot obtain high scores by hard-coding outputs or by only patching the public cases; they must implement robust parsing, core algorithms, and end-to-end artifact generation that generalize beyond the visible examples.
The official gold solution is required to achieve a verifier score of $1.0$ on the full hidden evaluation suite.

\paragraph{Difficulty calibration and validation.} We calibrate difficulty by repeatedly running Deepseek-V4-Pro~\cite{deepseekai2026deepseekv4} under 1.5-hour time budgets and adjusting task design until the tasks are challenging but still solvable in principle.
We generated 120 candidate tasks with quality filtering, where the final benchmark contains 46 tasks implemented in the Harbor format~\cite{merrill2026terminalbenchbenchmarkingagentshard}, all with containerized environments and a shared agent harness.

\subsection{Task composition}

The 46 tasks in \ours{} span 21 high-level categories, including interactive games, multimodal audits, software engineering, systems and performance, experiment reproduction, earth and environment, scientific computing, robotics, materials science, health and medicine, climate science, and chip design (EDA).
Figure~\ref{fig:lhtb-categories} shows the distribution.
No single category dominates the benchmark: the largest groups are interactive games and multimodal audits, with the remaining tasks spread across a broad range of scientific and engineering workflows.
% \section{Method}
% \label{sec:method}

% % Describe your approach. Use subsections for:
% %   - Overall architecture / pipeline
% %   - Key algorithms or design choices
% %   - Any theoretical insights or guarantees

% \subsection{Overall Approach}

% \subsection{Core Components}

% \subsection{Implementation Details}

\section{Experiments}
\label{sec:experiments}

\paragraph{Harness and agents.}
We evaluate models using the Harbor framework with the Terminus-2 agent harness~\cite{Harbor_Framework}, which interacts with each task environment through a single long-horizon terminal session. 
Additionally, for GPT-5.3, we instead use Codex~\cite{openai2025gpt5}\footnote{\url{https://github.com/openai/codex}} as the agent harness.
These harness frameworks allow us to evaluate how existing frontier models perform on sustained terminal-based workflows.

\paragraph{Models.}
We benchmark GPT-5.6-sol~\cite{openai2025gpt5}, GPT-5.5~\cite{openai2025gpt5}, GPT-5.4, GPT-5.3 Codex, DeepSeek V4 Pro~\cite{deepseekai2026deepseekv4}, Gemini 3.1 Pro~\cite{google2026gemini31pro}, GLM 5.1~\cite{glm5team2026glm5vibecodingagentic}, GLM 5.2, Kimi K2.6~\cite{moonshot2026kimik26}, Kimi K2.7 Code~\cite{moonshot2026kimik27code}, MiniMax M3~\cite{minimax2026minimaxm3}, Qwen3.7 Max~\cite{qwen2026qwen37}, Qwen3.6 Plus~\cite{qwen2026qwen36}, Doubao Seed 2.1 Pro~\cite{bytedance2026seed21}, Hy3~\cite{tencent2026hunyuan3}, Grok 4.20~\cite{xai2026grok420}, and Grok 4.5~\cite{xai2026grok45}.

% \subsection{Experimental Setup}

% \paragraph{Benchmark.}
% We evaluate on the 46-task Long-Horizon-Terminal-Bench (LHTB): the original 35 long-horizon tasks, six additive tasks (three useful, three non-useful), and five new multimodal audits (scientific-figure reconstruction, satellite flood change detection, microscopy cell-count QC, document table layout reconstruction, and audio-visual event alignment).
% The tasks span 21 domains, from interactive games and software engineering to chip design, climate science, and multimodal auditing (Figure~\ref{fig:lhtb-categories}).

\paragraph{Metrics.}
For each model we report pass@1 (fraction of tasks whose normalized reward is at least $0.95$), mean normalized reward across all tasks, number of episodes to complete the task, length of time, and estimated dollar cost and token usage per task.

\subsection{Main Results}

\begin{figure}[H]
  \centering
  \includegraphics[width=\linewidth]{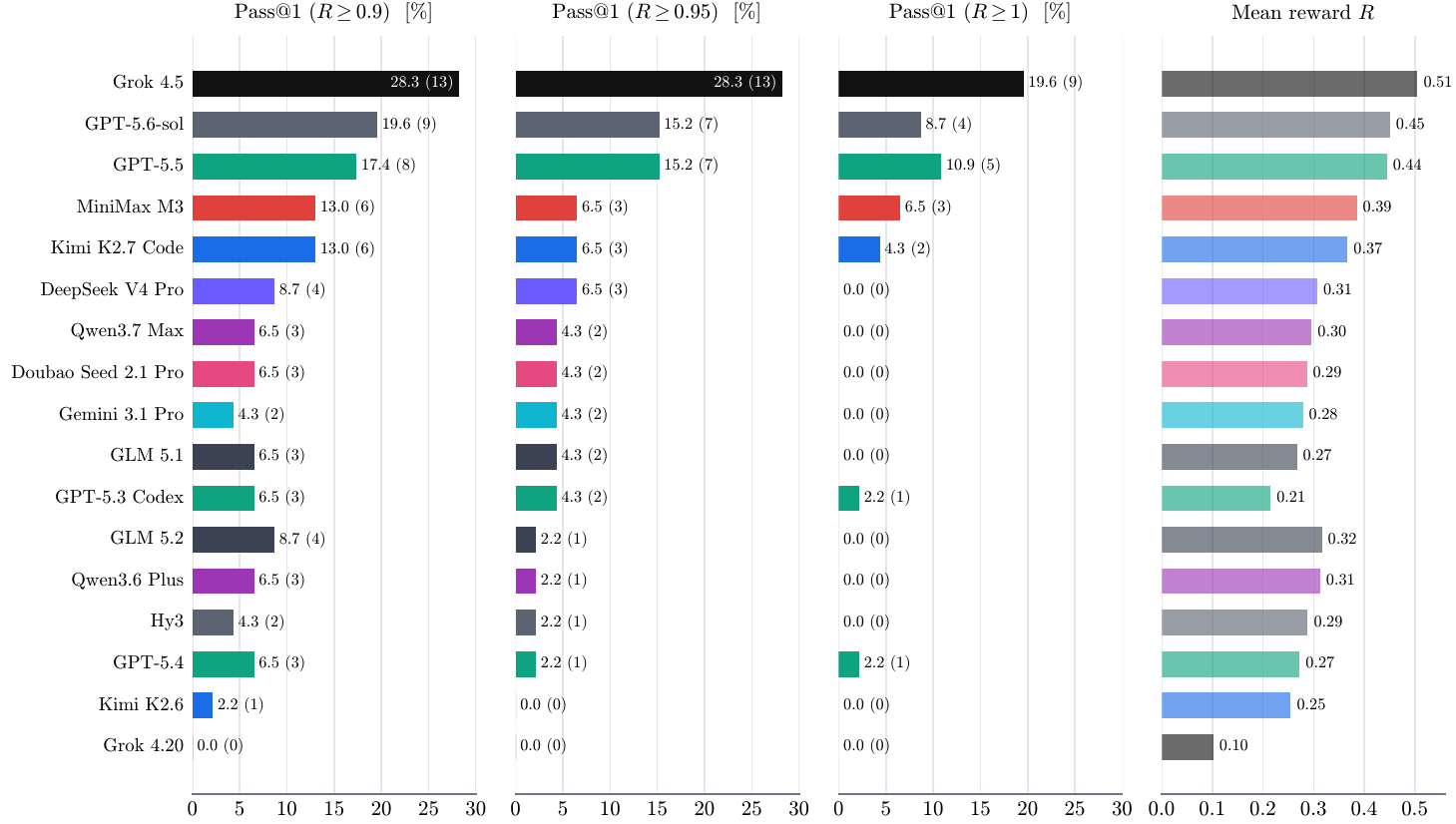}
    \caption{\textbf{LHTB leaderboard.}
    Pass@1 at reward thresholds $0.9$, $0.95$, and $1.0$, together with mean normalized reward, for each model under a shared agent setup, shown in a single model order sorted by pass@1 at $R \geq 0.95$.
Pass rate is the fraction of the 46 tasks whose overall reward $R$ meets a threshold $\tau$, with $R \geq 1.0$ corresponding to full completion, while mean $R$ captures partial progress across all tasks.
Grok 4.5 achieves the strongest overall performance, resolving $28.3\%$ of tasks at $R \geq 0.95$.
\textbf{Tightening the threshold to $R \geq 1.0$ collapses most models to zero pass rate, showing that partial-credit evaluation captures progress that a binary threshold misses.}}
  \label{fig:lhtb-leaderboard}
\end{figure}

Figure~\ref{fig:lhtb-leaderboard} summarizes pass rate at $R\geq0.9$, $R\geq0.95$, and $R\geq1.0$, along with mean reward across all 46 tasks.
Overall, the evaluated frontier models still struggle on long-horizon tasks, even under a dense-grading evaluation that gives credit for partial progress.
Grok 4.5 is the strongest reported model, but still resolves only $28.3\%$ of tasks (13/46) at $R\geq0.95$.
GPT-5.6-sol and GPT-5.5 follow at $15.2\%$ (7/46) each, with GPT-5.6-sol posting the slightly higher mean reward.
MiniMax M3, Kimi K2.7 Code, and DeepSeek V4 Pro follow at $6.5\%$ (3/46) each on the full 46-task run.
Below is a mid tier at $4.3\%$ that includes Qwen3.7 Max, Doubao Seed 2.1 Pro, Gemini~3.1~Pro, GLM~5.1, and GPT-5.3 Codex, while GLM~5.2, Qwen3.6 Plus, GPT-5.4, and Hy3 each resolve only a single task.
Kimi~K2.6 and Grok~4.20 solve zero tasks at $R\geq0.95$, with Grok~4.20 also recording the lowest mean reward ($R=0.10$) among the reported models.
Mean normalized rewards follow a similar ordering: Grok~4.5 also leads this metric at $R=0.51$, indicating that many models make partial progress on hard tasks without fully resolving them.
The results suggest that long-horizon execution is not just local reasoning task, but remains a bottleneck even for top frontier models.

\subsection{Dense Reward Is Necessary to Rank Models}
\label{sec:analysis-partial-credit}

\begin{figure}[H]
  \centering
  \includegraphics[width=0.92\linewidth]{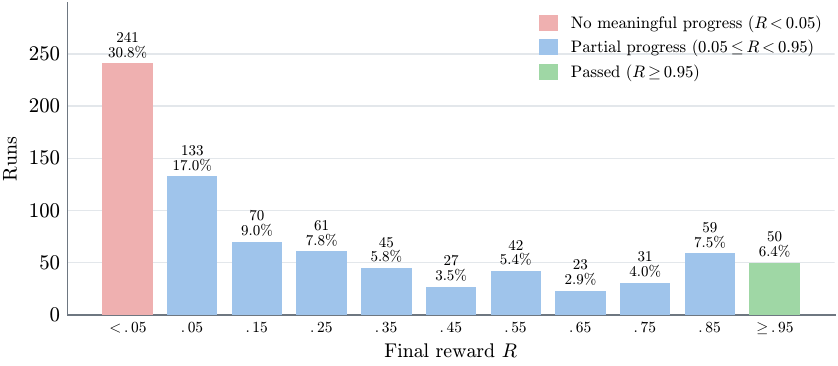}
\caption{\textbf{Distribution of final rewards over all $17 \times 46$ model--task runs.}
Only $50$ runs ($6.4\%$, green) pass the $R \geq 0.95$ threshold, while $241$ runs ($30.8\%$, gray) make no meaningful progress ($R < 0.05$).
The remaining $491$ runs ($62.8\%$, blue) achieve partial reward but would all be counted as failures under binary pass/fail evaluation.
This partial progress is often substantial rather than trivial: $223$ runs ($28.5\%$) reach $R \geq 0.5$, and the mass in $[0.85, 0.95)$ (68 runs) is still larger than the number of full passes.
\textbf{Binary grading collapses a wide range of partially successful runs into a single failure bucket, whereas dense subtask-level rewards preserve the signal needed to distinguish model progress at this difficulty level.}}
  \label{fig:lhtb-reward-dist}
\end{figure}

A key contribution of \ours{} is dense, subtask-level grading rather than binary pass/fail, which is essential at this difficulty level.
Under a strict threshold ($R \geq 1.0$), 10 of the 17 models pass zero tasks (Figure~\ref{fig:lhtb-leaderboard}); even at $R \geq 0.95$, only $6.4\%$ of runs pass while $30.8\%$ score below $0.05$ (Figure~\ref{fig:lhtb-reward-dist}).
The remaining $62.8\%$ of runs make real partial progress, but binary grading would count them as complete failures by giving them the same score as a run that makes no progress.

\paragraph{Binary grading hides meaningful differences between models.}
Under binary pass/fail evaluation, the leaderboard still collapses into large tie groups; for example, ten models each solve zero tasks at the strict $R \geq 1.0$ threshold.
Mean reward resolves these ties by quantifying how far a model progresses on tasks it does not fully complete.
Among the models tied at one solved task at $R\geq0.95$, mean reward spans from $0.27$ for GPT-5.4 to $0.32$ for GLM~5.2.
Pass rate and mean reward are positively but only moderately correlated (Spearman $\rho = 0.74$), and they can produce different model rankings because they measure different aspects of long-horizon performance: pass rate captures full task completion, whereas mean reward captures sustained partial progress.

\paragraph{Near-misses reveal real progress that binary pass/fail evaluation would hide.}
Dense grading also reveals \emph{near-misses} ($0.75 \leq R < 0.95$) runs that are close to completion but fail some final checks.
These near-misses occur nearly twice as often as passes ($90$ vs.\ $50$), indicating that many runs get very close to fully solving the task, but fail at the last few steps or checks.
For example, Kimi~K2.6 solves zero tasks at the $R \geq 0.95$ threshold, yet has five near-misses and a mean reward of $0.25$.
Its strongest near-miss reaches $R = 0.94$ on \texttt{grammar-fuzz-coverage-hunt}, and it also comes close on \texttt{spot-scheduler-traces} ($R = 0.90$), \texttt{poc-exploit-craft} ($R = 0.89$), and \texttt{nbody-accel-iterative} ($R = 0.89$).
Under binary grading, it would appear indistinguishable from a model that makes no meaningful progress, even though it repeatedly approaches successful completion.
As models improve, such near-misses are likely to provide an important early signal of progress, often revealing more clear capability gains before they are reflected in full pass rates.

\subsection{Cost Analysis}
\begin{figure}[H]
  \centering
  \includegraphics[width=\linewidth]{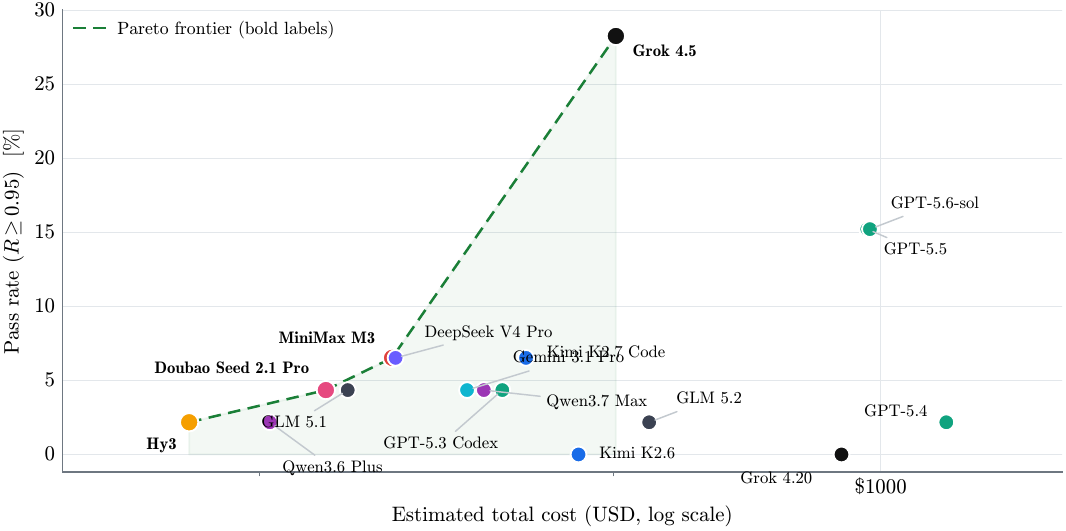}
  \caption{\textbf{Cost--reward frontier on LHTB.}
  Estimated total cost (log scale) against pass rate at $R\geq0.95$; the dashed line traces the Pareto frontier, and models on the frontier are shown in bold.
  Grok 4.5 achieves the highest pass rate among the reported models while remaining on the Pareto frontier, and Hy3 stays at the low-cost end of that frontier.}
  \label{fig:lhtb-cost-reward}
\end{figure}

Table~\ref{tab:lhtb-cost} reports the cost estimate from per-task token usage and public list prices, which Figure~\ref{fig:lhtb-cost-reward} shows the pass rate via the Pareto cost--reward frontier.
The average costs span from about \$3.6 to \$26 per task.
Although Grok 4.5 achieves the highest pass rate, it does so at only about \$11 per task, well below the most expensive models. GPT-5.6-sol and GPT-5.5 form a second tier of strong but substantially pricier models at roughly \$21 per task, and GPT-5.4 is the most expensive overall (about \$26 per task) with a much lower pass rate.
This gap arises because GPT-5.4 is the weaker model yet requires far more episodes per task (302 vs.\ 208 for GPT-5.5) at comparable per-M token pricing; it therefore has a higher cost while achieving a lower pass rate.

Hy3 anchors the low-cost end of the Pareto frontier at about $\$3.6$ per task, while Doubao Seed 2.1 Pro and MiniMax M3 also lie on the frontier, achieving pass@$R\geq0.95$ rates of $4.3\%$ and $6.5\%$ at roughly $\$5$ and $\$6$ per task.
Grok 4.5 extends that frontier upward, reaching $28.3\%$ pass@$R\geq0.95$ at roughly \$11 per task. GPT-5.6-sol matches GPT-5.5's pass rate at similar cost, but neither approaches Grok 4.5's cost-efficiency. DeepSeek V4 Pro sits just off the frontier, achieving the same $6.5\%$ pass@$R\geq0.95$ rate as MiniMax M3 at a similar cost. The remaining models, including Kimi K2.7 Code, GPT-5.3 Codex, GLM5.1, Gemini3.1 Pro, Qwen3.7 Max, GLM5.2, Kimi K2.6, and Grok 4.20, form a mid tier or lower tier with moderate to high costs but low resolution rates, indicating that higher inference spending alone is insufficient to guarantee better long-horizon performance.

\begin{table*}[t]
  \centering
  \small
  \begin{tabular}{lcccc}
    \toprule
    \textbf{Model} & \textbf{Tokens / task (M)} & \textbf{Episodes / task} & \textbf{Time / task (min)} & \textbf{Avg cost / task (\$)} \\
    \midrule
    Grok 4.5              &  8.91 & 197 & 74.9 & 10.95 \\
    GPT-5.6-sol           &  4.32 & 214 & 71.3 & 21.14 \\
    GPT-5.5               &  4.16 & 208 & 72.9 & 21.00 \\
    MiniMax M3            & \textbf{20.20} & 314 & 90.0 &  6.13 \\
    Kimi K2.7 Code        &  8.54 & 183 & 85.4 &  8.67 \\
    DeepSeek V4 Pro       & 14.45 & 321 & 83.6 &  6.19 \\
    Qwen3.7 Max           &  6.13 & 218 & 83.5 &  7.78 \\
    Doubao Seed 2.1 Pro   &  5.80 & 183 & 91.7 &  5.16 \\
    Gemini 3.1 Pro        &  3.55 & 148 & 85.0 &  7.44 \\
    GLM 5.1               &  5.84 & 120 & 92.6 &  5.46 \\
    GPT-5.3 Codex         &  4.99 & 305 & 84.6 &  8.16 \\
    GLM 5.2               &  8.43 & 195 & 89.3 & 11.93 \\
    Qwen3.6 Plus          &  8.67 & 194 & 88.6 &  4.47 \\
    Hy3                   & \textbf{25.33} & \textbf{435} & \textbf{176.1} &  3.62 \\
    GPT-5.4               & 10.90 & 302 & 79.3 & \textbf{25.77} \\
    Kimi K2.6             & 10.27 & 188 & 92.5 &  9.94 \\
    Grok 4.20             & 15.95 & 331 & 69.5 & 19.64 \\
    \midrule
    \textit{Average}      &  9.79 & 239 & 88.9 & 10.79 \\
    \bottomrule
  \end{tabular}
  \caption{\textbf{Estimated cost per model on LHTB.}
  On average, models spend 239 episodes and 88.9 minutes per task, at an estimated \$10.8 per task.
  Costs are estimated from per-task token usage and public list prices (USD per million tokens) as of June 2026.
  Tokens/task counts both input and output tokens; all prompt tokens are billed at the full input rate, since prompt-cache discounts are not guaranteed across providers.}
  \label{tab:lhtb-cost}
\end{table*}

% \subsection{Ablations and Analysis}

% \section{Failure Modes of Agents on Long-Hirizon Tasks}
% \label{sec:analysis}

% In this section, we do error analysis to discuss why current frontier models struggle on long-horizon terminal tasks.
% %
% We analyze the full run outputs, such as reward, termination cause, episode count, and wall-clock time for all $14 \times 46$ model task runs, and we identify four recurring patterns:
% \begin{enumerate}
%     \item A common failure mode is not taking obviously wrong actions but completing the tasks within a reasonable amount of time.
%     \item Even when agents terminate voluntarily before the timeout, they often do so prematurely, overestimating task completion despite still failing the hidden verifier. This points to stronger agent harnesses for self-verification and iterative debugging as an important direction for future work.
% \end{enumerate}

% \input{floats/fig_time_ablation}

\subsection{Dense Rewards Expose Different Failure Patterns}
\label{sec:analysis-timeout}

\begin{figure}[H]
  \centering
  \includegraphics[width=\linewidth]{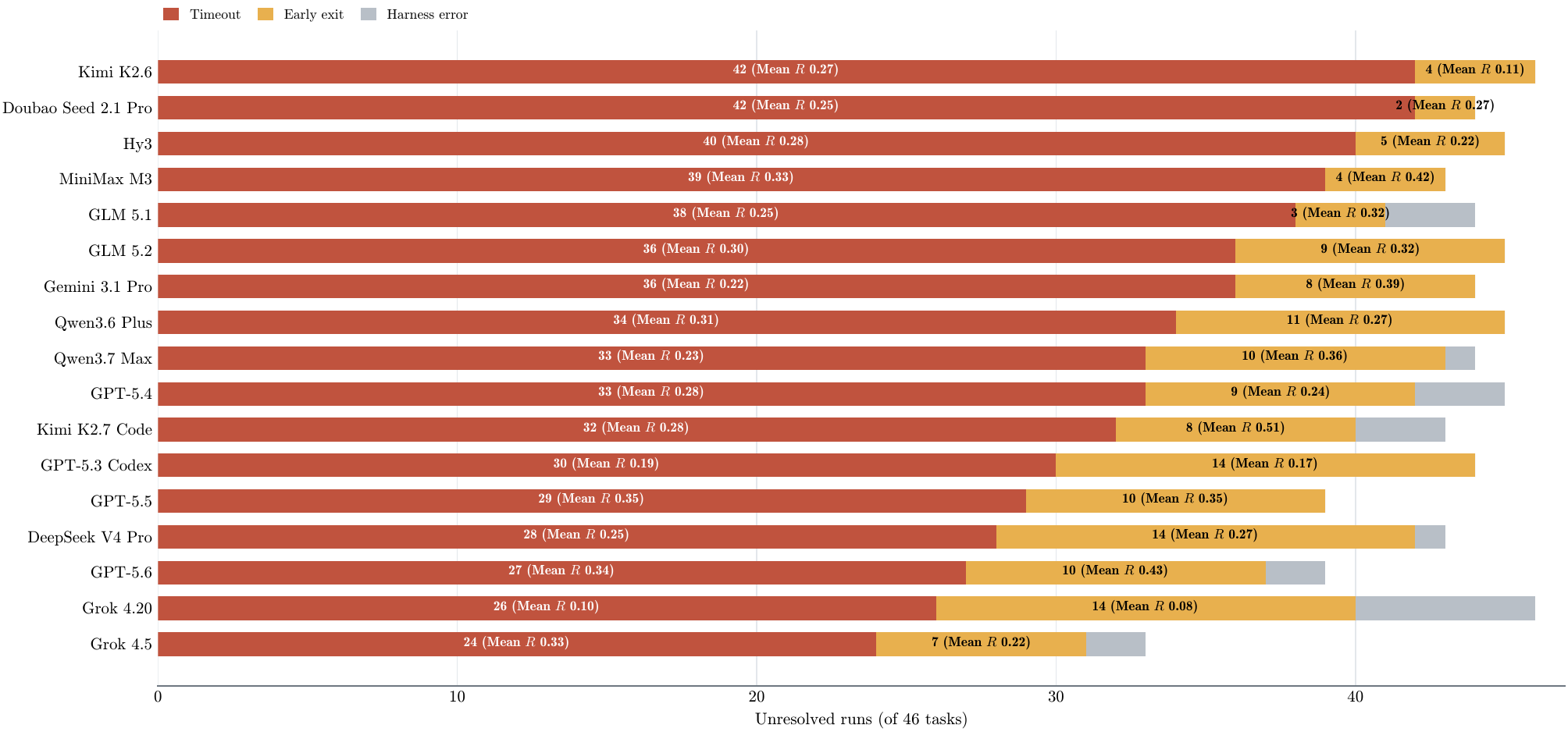}
  \caption{\textbf{Composition of unresolved runs per model.}
  Each bar decomposes a model's unresolved tasks ($R < 0.95$, out of 46) into timeouts (the agent is still working when the 90-minute budget expires), early exits (the agent terminates on its own with no harness error), and harness errors (non-timeout exceptions such as API- or verifier-side failures).
  The label inside each timeout segment shows the number of timed-out runs and their mean reward; the label inside each early-exit segment shows the number of early exits and their mean reward.
  Timeouts dominate for every model ($79\%$ of all unresolved runs), showing that many failures reflect incomplete progress rather than immediate breakdowns, while the smaller early-exit segments highlight cases where agents stop too soon despite not having fully solved the task.}
  \label{fig:lhtb-failure-modes}
\end{figure}

Figure~\ref{fig:lhtb-failure-modes} shows unresolved runs by termination cause.
Across all models, $79\%$ of unresolved runs ($518/660$) end because the 90-minute budget expires while the agent is still actively working.
However, these timed-out runs are not close to completion, with mean reward ranging only from 0.10 to 0.35 across models.
By contrast, early exits that account for only $19\%$ of unresolved runs often indicate overconfidence: the agent stops on its own despite not yet satisfying the hidden verifier.
Just $3\%$ are \emph{harness errors}, failures in the agent--environment loop caused by the evaluation harness itself, rather than by the model stopping early or exhausting its time budget.

\textbf{The main bottleneck is not local execution correctness, but long-horizon completion.}
On Terminal-Bench~2.0, most trials complete in under 20 minutes, and trajectory-level error analyses find that many failures are execution-related, including disobeying the specification, repeating steps, and missing termination conditions~\citep{merrill2026terminalbenchbenchmarkingagentshard}.
On \ours{}, the more common failure is different: agents can often string together many locally correct actions, but they cannot reliably turn that progress into a finished artifact before the horizon expires.
Short-horizon benchmarks primarily measure whether agents can \emph{act correctly}; \ours{} additionally measures whether they can \emph{budget a long horizon}, and the latter appears to be the binding constraint.

These results suggest that rankings on \ours{} reflect not only task-solving ability, but also \emph{time efficiency}: how much reward a model can accumulate within a fixed budget.
As a result, improvements that reduce redundant exploration, preserve state more effectively, or avoid repeated verification loops may yield larger gains than comparable improvements in single-step reasoning alone.

\subsection{Dense Rewards Reveal False Finishes and Weak Stopping Judgment}
\label{sec:analysis-false-finish}

Timeouts dominate model failures, but they are not the full story.
Among the 124 runs in which an agent terminates \emph{on its own} before the budget expires, a different limitation becomes clear: weak self-verification.

We define a \emph{false finish} as an early exit at relatively high reward, where the agent stops despite having not yet satisfied the hidden verifier.
We identify 14 such runs with $R \geq 0.75$.
In these cases, the agent has completed most of the task, judges itself finished, and exits with substantial time remaining.
For example, Kimi~K2.7~Code stops on \texttt{duckdb-optimizer-closure} at $R = 0.92$, GLM~5.2 stops on \texttt{apex-ib244-matter} at $R = 0.90$, and seven different models stop on \texttt{apex-law433-matter} between $R = 0.80$ and $0.87$, each with roughly 20 minutes still remaining.
These are not hopeless attempts, but near-complete ones in which the agent stops just short of satisfying the hidden verifier.

Dense rewards are essential for exposing this pattern.
A binary pass/fail metric would collapse all of these runs into the same failure bucket, whereas dense rewards show \emph{how far} each agent gets before stopping.
This reveals a clear axis of variation in stopping judgment (Figure~\ref{fig:lhtb-failure-modes}).
Kimi~K2.7~Code's early exits average $R = 0.51$ and MiniMax~M3's average $R = 0.42$, indicating that these models tend to stop later than other models ($R=0.22$ to $0.39$), after completing a substantial fraction of the task.
By contrast, Kimi~K2.6 exits at an average of only $R = 0.11$, abandoning tasks while most of the work remains.
Early exits reveal two distinct abilities: whether a model chooses to stop prematurely, and how accurately it can judge how much work remains before the task is truly complete.

This phenomenon is the long-horizon analogue of the \emph{verification} failures identified in Terminal-Bench~2.0, such as premature termination, weak verification, and missed final checks~\citep{merrill2026terminalbenchbenchmarkingagentshard}.
% Because \ours{} relies on hidden, artifact-based verifiers, agents cannot rely on their own internal summaries of progress.
After the obvious public errors are fixed, they must actively search for residual defects that no visible signal reveals.
The false-finish pattern suggests that current agents systematically overestimate completion and under-invest in final verification.
Closing this gap for long-horizon tasks will require not only stronger local reasoning, but also better self-verification and more calibrated stopping decisions.
\section{Related Work}
\label{sec:related}

% \paragraph{Terminal and software-engineering agent benchmarks.}
% Recent benchmarks have moved beyond static question answering toward realistic software and command-line work.
% SWE-Bench~\cite{jimenez2024swebench} and LiveCodeBench~\cite{jain2024livecodebench} evaluate repository-level bug fixing and code generation, while Terminal-Bench~\cite{merrill2026terminalbenchbenchmarkingagentshard} places agents in containerized terminals and asks them to complete high-skill technical tasks such as configuring systems, training models, and reproducing research code.
% More recent efforts push further toward difficult or prolonged software workflows: FrontierSWE~\cite{frontierswe2026} targets expert-level software engineering, and SWE-Marathon~\cite{desai2026swemarathon} studies ultra-long-horizon software work explicitly.
% These benchmarks have been instrumental in measuring agent progress, but most still emphasize a final solved/unsolved outcome and operate over horizons shorter than the workflows we target.
% \ours{} is closest in interface to Terminal-Bench, but differs in two key respects: its tasks are deliberately constructed to require sustained work over hundreds of steps and up to 90 minutes, and its subtask-based grading measures partial progress even when the final goal is not reached.

\paragraph{Terminal and software-engineering agent benchmarks.}
Recent agent benchmarks have shifted from static question answering and isolated programming exercises toward interactive, execution-grounded tasks in realistic work environments~\cite{zhou2023webarena,xie2024osworld}.
In software engineering, SWE-Bench~\cite{jimenez2024swebench} evaluates repository-level issue resolution on real GitHub projects, while LiveCodeBench~\cite{jain2024livecodebench} provides contamination-aware evaluation of code generation and related coding abilities such as execution, self-repair, and test-output prediction.
Subsequent benchmarks further broaden the scope of software-agent evaluation: SWT-Bench~\cite{mundler2024swt} studies test generation and validation for real-world bug fixes, and Terminal-Bench~\cite{merrill2026terminalbenchbenchmarkingagentshard} places agents in containerized terminals to complete realistic technical tasks such as system configuration, model training, and research-code reproduction.
More recent efforts push toward substantially harder and longer workflows, including FrontierSWE~\cite{frontierswe2026}, SWE-EVO~\cite{le2025swe}, SWE-Marathon~\cite{desai2026swemarathon}, and Edge-Bench, which target expert-level, multi-stage, or ultra-long-horizon tasks.
These benchmarks measure progress in coding and terminal agents, but many still primarily measure end-to-end success or failure, providing limited visibility into how much useful progress an agent made before failing.
\ours{} uses subtask-based grading exposes partial progress even when the final objective is not fully completed.

% \paragraph{Measuring long-horizon autonomy tasks.}
% Another closely related direction asks not only whether agents can solve a task, but how long a task they can solve reliably.
% METR's time-horizon analysis~\cite{kwa2025measuring} estimates the human-time duration of tasks that frontier agents can complete at a fixed success rate, and reports rapid growth in this quantity over time.
% This perspective highlights long-horizon autonomy as an increasingly important frontier, but also underscores the need for benchmarks that reveal more than a final success bit.
% \ours{} is designed for precisely this setting: its tasks require sustained, many-step execution, and its subtask-level grading reveals where along a long trajectory agents make progress or fail.

\paragraph{Measuring long-horizon autonomy tasks.}
Recent work has increasingly characterized long-horizon autonomy in terms of the duration of tasks that agents can complete reliably, rather than aggregate performance on a fixed benchmark~\cite{kwa2025measuring,wijk2024re}.
METR's time-horizon analysis~\cite{kwa2025measuring} makes this perspective explicit by estimating the human-time duration of tasks that frontier agents can complete at a fixed success rate.
This framing treats horizon length as a first-class capability variable and is grounded in human-calibrated evaluations such as RE-Bench~\cite{wijk2024re} and HCAST~\cite{rein2025hcast}, which compare agents against human performance on realistic research-engineering, software-engineering, cybersecurity, and reasoning tasks.
At the same time, it also highlights a limitation of outcome-only evaluation in long-horizon settings: a binary success signal provides little information about how much progress was made before failure or where execution began to diverge.
Recent studies show that long-horizon degradation can arise from the accumulation of execution errors over extended trajectories, even when short-horizon competence appears strong~\cite{sinha2025illusion}.
\ours{} is designed to complement this line of work by combining extended, many-step autonomy tasks with subtask-level grading, thereby enabling a finer-grained characterization of partial progress and failure throughout long-horizon execution.

% \paragraph{Beyond outcome-only grading.}
% Our grading design is also related to work on dense supervision and process-level rewards.
% Process reward models score intermediate reasoning steps rather than only final answers~\cite{lightman2023letsverify}, while rubric-based approaches decompose evaluation into individually checkable criteria and weighted dimensions of quality~\cite{gunjal2025rubrics,srar2026,tian2026arco}.
% Most of this literature uses partial-credit or step-level signals as \\emph{training} rewards, often with learned judges or LLM-based evaluators.
% By contrast, \ours{} brings the same intuition to \\emph{evaluation}: each task is decomposed into semantically meaningful subtasks, and each subtask is checked by deterministic, environment-grounded graders.
% This enables fine-grained analysis of long-horizon agent behavior without relying on a learned evaluator.

\paragraph{Beyond outcome-only grading.}
Process reward models provide supervision over intermediate reasoning steps rather than only final answers, yielding denser signals than outcome-only supervision~\cite{lightman2023letsverify,khalifa2025process}.
In parallel, rubric-based approaches decompose evaluation into individually assessable criteria or weighted dimensions of quality, making it possible to score complex outputs beyond a single holistic judgment~\cite{rao2026autorubric,pan2026rubriceval,tian2026arco}.
Most of this literature uses fine-grained signals primarily as \emph{training} rewards or judge-based supervision, often relying on learned verifiers or LLM evaluators~\cite{uesato2022solving,zheng2023judging,li2025self,yuan2024self}.
\ours{} applies the same core intuition at \emph{evaluation} time: each task is decomposed into semantically meaningful subtasks, and each subtask is checked by deterministic, environment-grounded graders.

% \paragraph{Agent Harnesses.}
% Benchmark results depend not only on the underlying model, but also on the scaffold through which it acts.
% Open systems such as SWE-agent~\cite{yang2024sweagent}, OpenHands~\cite{wang2024openhands}, and the Terminus family introduced alongside Terminal-Bench~\cite{merrill2026terminalbenchbenchmarkingagentshard} provide agent--computer interfaces for autonomous repository and terminal work.
% To reduce scaffold-induced variance and make cross-model comparisons cleaner, we evaluate all models in \ours{} under the same shared \texttt{terminus-2} agent.

\paragraph{Agent harnesses.}
Benchmark outcomes depend not only on the underlying model, but also on the agent harness that mediates its interaction with the environment~\cite{yao2026harness, yang2024sweagent,li2026comfyclaw,lin2026harness}. 
Open frameworks such as SWE-agent~\cite{yang2024sweagent}, OpenHands~\cite{wang2024openhands}, and the Terminus family introduced alongside Terminal-Bench~\cite{merrill2026terminalbenchbenchmarkingagentshard} provide reusable interfaces for autonomous repository- and terminal-based work. Opensource agents such as Codex~\cite{openai2025codex} and OpenClaw ~\cite{openclaw2026} package nontrivial scaffolding around the base model, including prompting, tool orchestration, context management, and action-selection policies. 

\section{Conclusion}
\label{sec:conclusion}

We introduced \ours{}, a benchmark of 46 containerized terminal tasks spanning nine domains and designed to stress long-horizon execution.
Unlike outcome-only benchmarks, \ours{} uses deterministic, environment-grounded subtasks to provide dense partial-credit signals, allowing evaluation to capture not only whether an agent finishes a task, but also how far it progresses.

Across 17 frontier models under a shared \texttt{terminus-2} agent, tasks in \ours{} require substantial effort, averaging 239 episodes, 9.8M tokens, \$10.8 in API cost, and 88.9 minutes per run.
Yet the benchmark remains far from saturated: the strongest model, Grok 4.5, achieves only a $28.3\%$ pass rate at $R \geq 0.95$, while the mean pass rate across models is just $6.4\%$.
Dense rewards further reveal that many runs make meaningful but incomplete progress, exposing failure patterns that binary pass/fail would miss.

Our analysis suggests that the main bottleneck is not only local reasoning, but reliable long-horizon completion.
Current agents often time out after making partial progress, and even voluntary early exits frequently reflect weak self-verification rather than true completion.
These results highlight the need for better planning, stronger memory and progress tracking, and more calibrated stopping decisions for long-horizon tasks.
We release \ours{} and its evaluation harness to support future work on agents that can plan, verify, and execute reliably over long horizons.

\bibliographystyle{plainnat}
\bibliography{refs}

\newpage
\appendix
\section{List of Tasks in \ours{}}
\label{app:task-list}

Table~\ref{tab:lhtb-task-list} lists all 46 benchmark tasks using the nine-category taxonomy used throughout the paper. The difficulty label is derived from the task's mean reward across model results Figure~\ref{fig:lhtb-leaderboard}: tasks with mean reward at least $0.5$ are labeled \emph{Easy}, and the rest are labeled \emph{Hard}.

\small
\setlength{\LTleft}{0pt}
\setlength{\LTright}{0pt}
\begin{longtable}{p{0.18\textwidth}p{0.20\textwidth}p{0.50\textwidth}p{0.07\textwidth}}
\caption{\textbf{Task list for \ours{}.} Each row shows a task ID, its paper-level category, a one-sentence description, and a coarse difficulty label derived from average task reward in Figure~\ref{fig:lhtb-leaderboard} (\emph{Easy} if mean reward $\geq 0.5$, otherwise \emph{Hard}).}
\label{tab:lhtb-task-list}\\
\toprule
\textbf{Task ID} & \textbf{Category} & \textbf{Description} & \textbf{Difficulty} \\
\midrule
\endfirsthead

\toprule
\textbf{Task ID} & \textbf{Category} & \textbf{Description} & \textbf{Difficulty} \\
\midrule
\endhead

\bottomrule
\endfoot

2048 & Interactive games & Play 2048 turn-by-turn via a terminal game server, merging tiles to reach the highest tile, scored by replaying recorded moves in a fresh seeded engine. & Hard \\
alp-paper-reproduction & Research reproduction \& ML & Reproduce a calibrated ALP floating-point compression experiment from the paper's method and sample columns. & Hard \\
apex-ib244-matter & APEX professional workflows & Advise on taking EdTech firm KSchool private across 38 reactive stages, rebuilding LBO/DCF/comps models for exact figures. & Easy \\
apex-investment-banking-matter & APEX professional workflows & Work a private-equity take-private of Planet Fitness through 19 reactive stages, rebuilding an LBO model and answering exact figures. & Hard \\
apex-law433-matter & APEX professional workflows & Act as outside counsel across four legal matters and 70 reactive stages over \textasciitilde{}600 documents, submitting deterministically-graded answers. & Easy \\
apex-management-consulting-matter & APEX professional workflows & Work two management-consulting client matters across 33 reactive stages, rebuilding business-case and cost-savings models for exact answers. & Hard \\
apex-openroad-ibex-signoff & Software \& reverse engineering & Close RTL-to-GDSII signoff on the ibex CPU with OpenROAD/sky130 across reactive stages until timing, DRC, LVS, area, and power pass. & Hard \\
audio-visual-event-alignment & Multimodal \& imaging analysis & Repair an audio-video sync audit that detects visual events, audio onsets, dropped frames, and per-clip synchronization offsets. & Hard \\
chess-mate & Interactive games & Play chess as White against a bundled superhuman Stockfish engine, scored on how long the agent survives before being checkmated. & Hard \\
climate-netcdf-extreme-event-audit & Earth, climate \& energy & Repair a climate audit pipeline computing gridded heatwave, precipitation, drought, anomaly, and index metrics over NetCDF datasets. & Hard \\
commit0-multilib-tdd & Software \& reverse engineering & Reimplement three whole Python libraries from stubs with no provided tests, writing your own suite, graded on hidden real test suites. & Hard \\
dicom-radiology-audit & Multimodal \& imaging analysis & Audit synthetic DICOM CT series geometry, HU conversion, windowing, masks, and lesion measurements. & Hard \\
document-table-layout-reconstruction & Multimodal \& imaging analysis & Repair a scanned-document pipeline that detects table grids, assigns noisy OCR tokens to cells, and emits CSV/JSON/Markdown tables. & Hard \\
duckdb-optimizer-closure & Systems, performance \& security & Iteratively patch DuckDB's query optimizer to speed up TPC-H while preserving correctness. & Hard \\
epa-swmm-stormwater-regression-audit & Earth, climate \& energy & Audit EPA SWMM stormwater hydraulic and hydrologic simulations against official solver outputs. & Hard \\
epidemic-inverse-control-audit & Scientific computing \& simulation & Fit and control a multi-city, age-structured SEIR-H-ICU epidemic model. & Hard \\
foldseek-paper-reproduction & Research reproduction \& ML & Reproduce a calibrated Foldseek-style structural protein search experiment from the Nature Biotechnology paper. & Hard \\
gdal-proj-raster-regression & Earth, climate \& energy & Reproduce GDAL/PROJ raster reprojection, coordinate transform, point sampling, and zonal-statistics regression checks. & Hard \\
generals-bot-arena & Software \& reverse engineering & Develop a bot for a Generals.io-style grid strategy game by iteratively playing matches and refining its strategy. & Easy \\
grammar-fuzz-coverage-hunt & Systems, performance \& security & Iteratively build a grammar-guided fuzzer to maximize code coverage of target parsers. & Easy \\
great-expectations-audit & Software \& reverse engineering & Build a Great Expectations-based multi-table data-quality and reconciliation audit pipeline. & Hard \\
langchain-version-migration & Software \& reverse engineering & Migrate a legacy LangChain RAG/router app across a breaking API change with hard-fail compatibility gates. & Hard \\
materials-phase-diagram-audit & Scientific computing \& simulation & Repair a materials phase-diagram audit reconstructing phase boundaries, invariant reactions, lever-rule fractions, and convex-hull stability. & Hard \\
matpower-opf-regression & Earth, climate \& energy & Reproduce MATPOWER/PYPOWER power-flow and optimal-power-flow regression checks with grid-constraint diagnostics. & Hard \\
microscopy-cell-count-qc-audit & Multimodal \& imaging analysis & Repair a microscopy QC audit that segments synthetic cell imagery, flags focus/artifact failures, and reports per-tile channel counts. & Hard \\
modflow6-groundwater-regression-audit & Earth, climate \& energy & Audit USGS MODFLOW 6 groundwater simulations against official solver outputs. & Hard \\
nbody-accel-iterative & Scientific computing \& simulation & Iteratively optimize an N-body gravitational simulation from O(N\textasciicircum{}2) to O(N log N) with correctness verification at multiple scales. & Easy \\
nrel-pysam-hybrid-renewables-audit & Earth, climate \& energy & Audit renewable generation and storage dispatch against NREL PySAM official PVWatts outputs. & Hard \\
opensees-seismic-structural-regression-audit & Scientific computing \& simulation & Audit seismic structural-dynamics results against official OpenSeesPy finite-element solver outputs. & Hard \\
poc-exploit-craft & Systems, performance \& security & Generate proof-of-concept inputs that trigger specific vulnerability classes using sanitizer feedback. & Easy \\
riscv-core-debug & Software \& reverse engineering & Localize and fix undisclosed injected bugs in an out-of-order RISC-V SystemVerilog CPU until simulation, compliance, and formal checks pass. & Easy \\
robotics-slam-benchmark-repair & Research reproduction \& ML & Repair a SLAM benchmark audit that composes SE(2) poses, evaluates loop closures, reconstructs landmark maps, and reports trajectory consistency. & Hard \\
rush\_hour\_campaign & Logic \& constraint puzzles & Solve a four-stage 6x6 Rush Hour campaign by hand, submitting legal move routes under length caps. & Hard \\
satellite-flood-change-detection-audit & Multimodal \& imaging analysis & Repair a satellite flood change-detection audit that compares before/after imagery, masks clouds, and reports per-tile flood extent. & Hard \\
scientific-figure-data-reconstruction & Multimodal \& imaging analysis & Repair a pipeline that reconstructs plotted series data from scientific figure images, captions, and partial CSV references. & Hard \\
snake\_maze\_campaign & Interactive games & Play a deterministic hard Snake variant, eating seeded apples and surviving growing obstacles as long as possible, scored by replaying the move log. & Hard \\
sokoban & Logic \& constraint puzzles & Solve a ramp of Sokoban puzzles turn-by-turn through a game server, advancing as far as possible, scored by replaying moves in a clean engine. & Hard \\
spice-ephemeris-regression & Scientific computing \& simulation & Reproduce NASA SPICE ephemeris, angle, and ground-station visibility regression checks from pinned kernels. & Hard \\
spot-scheduler-traces & Systems, performance \& security & Implement and refine a cloud spot-instance scheduling policy that minimizes cost while meeting hard deadlines. & Easy \\
su2-airfoil-regression & Scientific computing \& simulation & Reproduce a hardened SU2 aerodynamic regression matrix from official test-case references and solver history artifacts. & Hard \\
sudoku-recovery & Logic \& constraint puzzles & Recover seven progressively corrupted hard variant-Sudoku puzzles turn-by-turn, detecting subtle corrupt givens via global constraint reasoning. & Hard \\
super-mario & Interactive games & Play Super Mario Bros turn-by-turn through a game server, clearing as many levels as possible, scored by replaying frame inputs in a fresh emulator. & Hard \\
tabular-data-feature-covshift & Research reproduction \& ML & Recover a sparse high-dimensional signal from low-dimensional observations under covariate shift, with a leakage-audited honesty gate. & Hard \\
unison-paper-reproduction & Research reproduction \& ML & Reproduce a calibrated UNISON fat-tree network simulation experiment from the paper's method and settings. & Hard \\
unknown-config-semantics & Software \& reverse engineering & Reverse-engineer an undocumented config format from a noisy spec and repair its normalization engine to pass hidden probes across five stages. & Hard \\
vector-db-iterative-build & Systems, performance \& security & Build an approximate nearest-neighbor vector search service from scratch, iteratively optimizing recall and throughput. & Hard \\
\end{longtable}
\normalsize

\end{document}